\documentclass[11pt,a4paper]{article}
\usepackage[hyperref]{naaclhlt2019}
\usepackage{times}
\usepackage{latexsym}

\usepackage{graphicx}
\usepackage{amsfonts}
\usepackage{amsmath}
\usepackage{url}
\usepackage{verbatim}
\usepackage{booktabs}
\usepackage{tabularx}
\usepackage{rotating}
\usepackage{soul}
\usepackage{xcolor}
\usepackage[normalem]{ulem}
\usepackage{cleveref}

\usepackage{etoolbox}
\makeatletter
\patchcmd\@combinedblfloats{\box\@outputbox}{\unvbox\@outputbox}{}{\errmessage{\noexpand patch failed}}
\makeatother

\newcommand{\ctext}[3][RGB]{%
  \begingroup
  \definecolor{hlcolor}{#1}{#2}\sethlcolor{hlcolor}%
  \hl{#3}%
  \endgroup
}
\newcommand{\tabsec}[1]{ \multicolumn{1}{c}{\textbf{#1}} \\ }
\newcommand{\hlArt}[1]{ \ctext[RGB]{102,214,67}{#1} }
\newcommand{\hlCtx}[1]{ \ctext[RGB]{249,224,71}{#1} }
\newcommand{\hlRef}[1]{ \ctext[RGB]{253,146,38}{#1} }
\newcommand{\confint}[1]{{\small{}\textcolor{gray}{ (#1)}}}

\aclfinalcopy 
\def\aclpaperid{1995} 

\title{Towards Content Transfer through Grounded Text Generation}

\author{Shrimai Prabhumoye \\
  Carnegie Mellon University \\
  5000 Forbes Avenue \\
  Pittsburgh, PA 15219 \\
  {\tt sprabhum@andrew.cmu.edu} \\\And
  Chris Quirk, Michel Galley\\
  Microsoft Research \\
  One Microsoft Way \\
  Redmond, WA 98052 \\
  {\tt \{chrisq,mgalley\}@microsoft.com}}

\date{}

\begin{document}
\maketitle
\begin{abstract}
Recent work in neural generation has attracted significant interest in controlling the {\it form} of text, such as style, persona, and politeness. 
However, there has been less work on controlling neural text generation for content.
This paper introduces the notion of Content Transfer for long-form text generation, where the task is to generate a next sentence in a document that both fits its context and is grounded in a content-rich external textual source such as a news story. Our experiments on Wikipedia data show significant improvements against competitive baselines. As another contribution of this paper, we release a benchmark dataset of 640k Wikipedia referenced sentences paired with the source articles to encourage exploration of this new task.
\end{abstract}

\section{Introduction}

\begin{figure}[ht]
\centering
\includegraphics[width=3.0in]{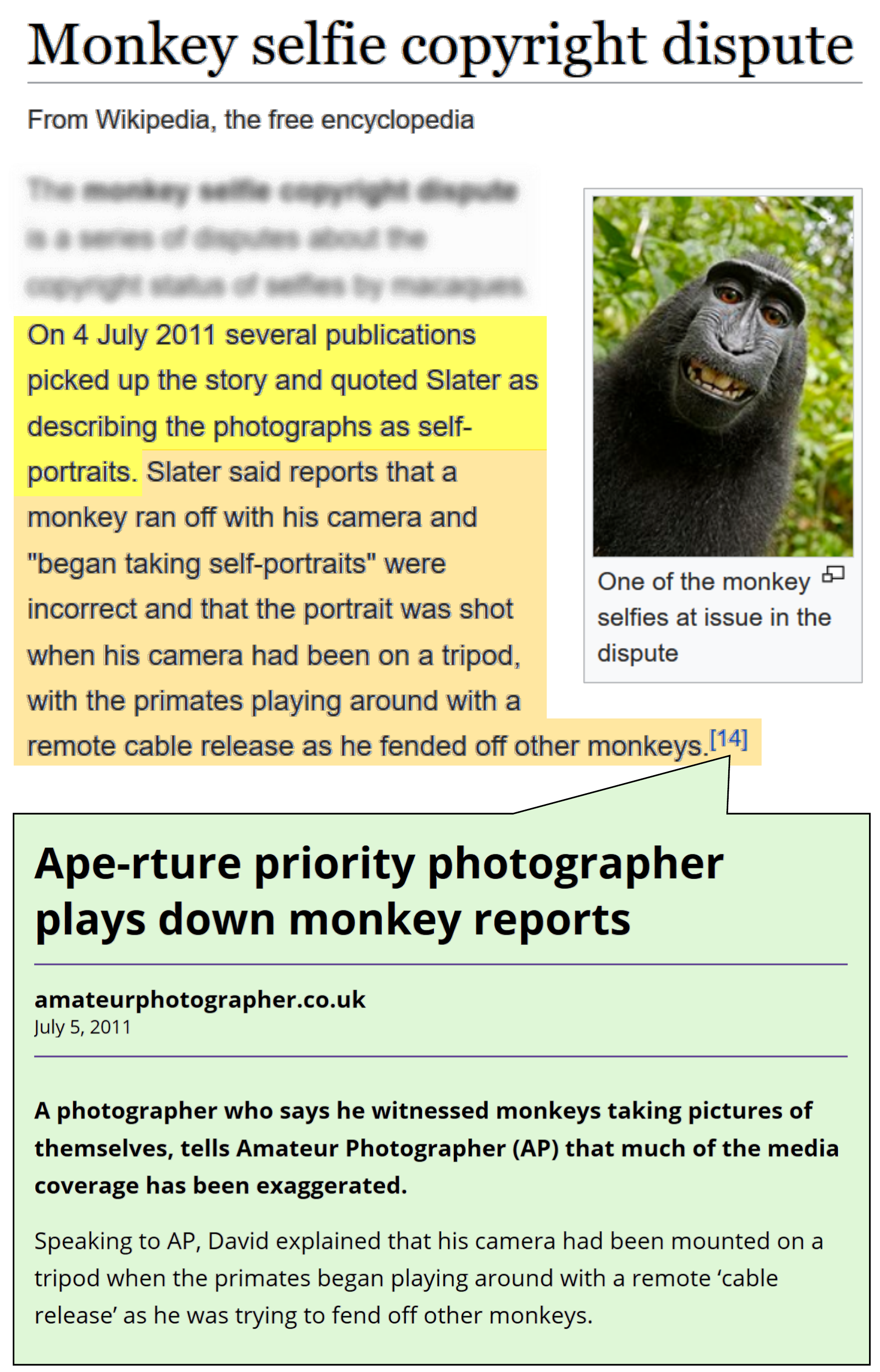}
\vspace{-1em}
\caption{
Example of content transfer:
Given existing curated text (yellow) and a document with additional relevant information (green),
the task is to update the curated text (orange) to reflect the most salient updates.
}
\label{fig:selfie}
\vspace{-1em}
\end{figure}

Recent work in neural natural language generation (NLG) has witnessed a growing interest in controlling text for various form-related and linguistic properties, such as
style \cite{ficler:2017},
affect \cite{ghosh:2017},
politeness \cite{sennrich:2016}, persona \cite{li:2016} voice \cite{yamagishi:2016},
grammatical correctness \cite{ji:2017}, and
length \cite{kikuchi:2016}.
This trend offers the promise of empowering existing authoring tools such as Grammarly, Google Smart Compose, and Microsoft Word with the ability to control a much greater variety of textual properties,
which are currently mostly limited to grammar, spelling, word choice, and wordiness.
What has been relatively less explored in neural NLG research is the ability to control the generation of a current sentence not only in its {\it form}, but also its {\it content}.\footnote{Historically, NLG has focused on generation from structured content such as a database or semantic representation, but this paper is interested in generation from free-form text.}
Consider for example Fig.~\ref{fig:selfie}, which illustrates a situation where an author edits a document (here a Wikipedia article), and the goal is to generate or suggest a next sentence (shown in orange) to the author. This type of unconstrained, long-form text generation task \cite{mostafazadeh:2016,fan:2018} is of course extremely difficult.
Free-form generation can easily go astray due to two opposing factors. 
On one hand, ensuring that the generated output  
is of relatively good quality
often comes at the cost of making it bland and devoid of factual content \cite{li:2015}. On the other hand, existing techniques can help steer neural models away from blandness in order to produce more contentful outputs (using temperature sampling \cite{fan:2018}, GAN \cite{gan:2014}, etc.), but often at the cost of ``hallucinating'' \cite{wiseman:2017} words or concepts that are totally irrelevant. Neither situation provides a compelling experience to the user.

What is clearly missing from the aforementioned authoring scenario is the notion of {\it grounding}: there is often a profusion of online resources that bear at least some relevance to any given document currently being written.
Much of the general-purpose world knowledge is available in the form of encyclopedias (e.g., Wikipedia), 
books 
(e.g., Project Gutenberg, Google Books), and news articles.
While the generation of good quality texts without any conditioning on ``external'' sources \cite{fan:2018} might be an interesting research endeavor on its own, we argue that grounding can make the generation task much easier, e.g., as shown in Fig.~\ref{fig:selfie} where a passage of a news article (green) can be reformulated considering the current context of the document (yellow) in order to produce a natural next sentence (orange).
In light of this desideratum, this paper addresses the problem of grounded text generation, where the goal is to infuse the content or knowledge from an external source (e.g., a news article as in Fig.~\ref{fig:selfie}) in order to generate a follow-up sentence of an existing document. We see this as a form of {\it Content Transfer}, as other characteristics of the external source---such as style and linguistic form---are not controlled.

In addition to formulating this new task, our work makes the following contributions:
We provide a large dataset of 640k instances that contain parallel data of a source document (news articles), a context, and sentence to be produced. 
The latter two are extracted from Wikipedia, which is an attractive dataset for grounded generation as many of the statements in Wikipedia cite external sources (i.e., grounded in an external article).
Finally, we also provide simple yet efficient models that condition both on the external article and the context of the current document. 
We compare our models against extractive and abstractive baselines, including summarization methods that simply try to condense the external article without considering the context of the document. 
Our experiments show that our models which incorporate the context gain 7.0 ROUGE-L F1 points 
---in other words, treating our task as a summarization problem is not enough.
Our human evaluations also show that models that are aware of the context generate relevant and fluent sentences that are coherent to the context.

\section{Task}
\label{sec:content_transfer}

This research is concerned with the general problem of {grounded authorship assistance}, 
i.e., the task of suggesting text to insert or append in an existing document draft, in such a way that all the added {\it content} reflects information from external sources, such as news articles and books. 
This type of grounded generation task could take many forms, so we decided to formalize the task as follows, while still keeping the task both challenging and practically interesting. 
Given an external {\bf document} (green in Fig.~\ref{fig:selfie}), and some existing {\bf curated text} (yellow), the task is to generate a single {\bf update sentence} (orange).
This update sentence should be both relevant to the context and reflective of the information contained in the document. 

This task bears some similarity with automatic summarization \cite{nenkova:2011}, as a na\"ive approach to the above problem is to append a one-sentence summary of the document to the curated text. 
While indeed related, the two tasks differ in two key points. 
First, the one-sentence summary must be contextually appropriate given the previous context of the curated text. 
Second, summarization is mostly concerned with finding {\it salient} information, but---in the case of our task---information relevant to the context might actually only be auxiliary within the external document. 
Section~\ref{sec:related} (Related Work) further contrasts our task with summarization.

Formally we define our task as follows: given an existing curated text $\boldsymbol{s}$ and a document $\boldsymbol{d}$ describing novel information relevant to that text, the system must produce a revised text $\boldsymbol{s'}$ that incorporates the most salient information from $\boldsymbol{d}$.
We restrict our focus to the cases where the revised text $\boldsymbol{s'}$ can be obtained by appending the new information from $\boldsymbol{d}$  to the original curated text $\boldsymbol{s}$.\footnote{
In general, updated information from $\boldsymbol{d}$ might demand substantial changes to $\boldsymbol{s}$: perhaps core assumptions of $\boldsymbol{s}$ were contradicted, necessitating many removed and rewritten sentences.
We postpone this complex setting to future work.
}
In particular, we assume we can transform the old curated text $\boldsymbol{s}$ into the new text $\boldsymbol{s}'$ by appending one additional update sentence $\boldsymbol{x}$ to $\boldsymbol{s}$.

\section{Models}

This paper operates in a conventional supervised learning setting.
For training data, we rely on a large dataset of existing curated text $\boldsymbol{S} = \{\boldsymbol{s}_{1},\ldots,\boldsymbol{s}_{n}\}$, corresponding documents with novel information $\boldsymbol{D} = \{\boldsymbol{d}_{1},\ldots,\boldsymbol{d}_{n}\}$, and the update sentences $\boldsymbol{X} = \{\boldsymbol{x}_{1},\ldots,\boldsymbol{x}_{n}\}$.
Our task is to generate the update sentence $\boldsymbol{x}_i$ that could be appended to the curated text $\boldsymbol{s}_i$ in order to incorporate the additional information from document $\boldsymbol{d}_i$. 
The goal would be to identify new information (in particular, $\boldsymbol{d}_i \setminus \boldsymbol{s}_i$) that is most salient to the topic or focus of the text, then generate a single sentence that represents this information.

\subsection{Generative models}

A natural though difficult means of generating this additional update sentence $\boldsymbol{x}$ is to use a generative model conditioned on the information in the curated text $\boldsymbol{s}$ and the new document $\boldsymbol{d}$.
Recent methods inspired by successful neural machine translation systems have produced impressive results in abstractive summarization \cite{nallapati2016abstractive}.
Hence, our first step is to use the sequence-to-sequence encoder-decoder model \cite{bahdanau2015neural} with attention \cite{DBLP:journals/corr/LuongPM15} for our task.
This kind of model assumes that the output sentence can be generated word-by-word.
Each output word $\boldsymbol{x}_i^t$ generated is conditioned on all prior words $\boldsymbol{x}_i^{<t}$ and an encoded representation of the context $\boldsymbol{z}$:
\begin{equation}
    \prod_{t} p(\hat{\boldsymbol{x}}_{i}^{t}| \hat{\boldsymbol{x}}_{i}^{<t}, \boldsymbol{z})
\end{equation}

\paragraph{Context Agnostic Generative (CAG) Model:} 
One simple baseline is to train a sequence-to-sequence model for the document $\boldsymbol{d}$ alone that does not directly incorporate information from the curated text $\boldsymbol{s}$.
Here, the algorithm is trained to generate the most likely update sentence $\hat{\boldsymbol{x}} = \arg\max p(\boldsymbol{x} | \boldsymbol{d})$.
In this setting, we consider the reference document $\boldsymbol{d}_{i}$ as the source and the update sentence to be generated $\boldsymbol{x}_{i}$ as the target. 

\begin{equation}
    \boldsymbol{z} = \text{Encoder}(\boldsymbol{d}_{i}, \boldsymbol{\theta})
\end{equation}

The encoder and decoder do not directly see the information from the curated text $\boldsymbol{s}$, but the update $\boldsymbol{x}$ inherently carries some information about it.
The parameters of the model are learned from updates that were authored given the knowledge of the curated text.
Hence, the model may capture some generalizations about the kinds of information and locations in $\boldsymbol{d}$ that are most likely to contribute novel information to $\boldsymbol{s}$. 

\paragraph{Context Only Generative (COG) Model:} 
This algorithm is trained to generate the most likely update sentence $\hat{\boldsymbol{x}} = \arg\max p(\boldsymbol{x} | \boldsymbol{s})$.
This model is similar to CAG except that we consider the curated $\boldsymbol{s}_{i}$ as the source.
In this setting, there is no grounding of the content to be generated.

\paragraph{Context Informed Generative (CIG) Model:} 
An obvious next step is to incorporate information from the curated text $\boldsymbol{s}$ as well.
We can concatenate the document and the curated text, and produce an encoded representation of this sequence.
\begin{equation}
    \boldsymbol{z} = \text{Encoder}([\boldsymbol{d}_{i}; \boldsymbol{s}_{i}], \boldsymbol{\theta})
\end{equation}
This approach incorporates information from both sources, though it does not differentiate them clearly.
Thus, the model may struggle to identify which pieces of information are novel with respect to the curated text.

To clearly identify the information that is already present in the curated text $\boldsymbol{s}$, a model could encode $\boldsymbol{s}$ and $\boldsymbol{d}$ separately, then incorporate both signals into the generative procedure.

\paragraph{Context Receptive Generative (CRG) Model:}
Our next step was to condition our generative process more concretely on the curated text $\boldsymbol{s}$. 
We condition the generative process on the representation of $\boldsymbol{s}$ at each time step. 
Formally:

\begin{eqnarray}
    \boldsymbol{z_d} &=& \text{Encoder}_{\boldsymbol{d}}(\boldsymbol{d}_{i}, \boldsymbol{\theta_{d}}) \\
    \boldsymbol{z_s} &=& \text{Encoder}_{\boldsymbol{s}}(\boldsymbol{s}_{i}, \boldsymbol{\theta_{s}}) \\
    \boldsymbol{\hat{x}_{i}} &\sim& \prod_{t} p(\hat{x}_{i}^{t} |  [\hat{\boldsymbol{x}}_{i}^{<t}; \boldsymbol{z_{s}}], \boldsymbol{z_{d}})
\end{eqnarray}
where, $\boldsymbol{\theta_{d}}$ and $\boldsymbol{\theta_{s}}$ are the parameters of the encoder for the document $\boldsymbol{d}$ and encoder for the curated text $\boldsymbol{s}$ respectively, $\boldsymbol{z_{d}}$ and $\boldsymbol{z_{s}}$ are the encoded representations of the document $\boldsymbol{d_{i}}$ and curated text $\boldsymbol{s_{i}}$  respectively. 
At each time step of generation, the output is conditioned on the tokens generated up to the time step $t$ concatenated with $\boldsymbol{z_{s}}$.
Hence, the generative process is receptive of the context at each time step.

\subsection{Extractive models}

Generative models that construct new sentences conditioned on the relevant context are compelling but have a number of modeling challenges.
Such a model must both select the most relevant content \emph{and} generate a fluent linguistic realization of this information.

We also consider extractive models: approaches that select the most relevant sentence from the document $\boldsymbol{d}$ to append to the curated text $\boldsymbol{s}$.
These approaches can focus solely on the content selection problem and ignore the difficulties of generation.
This simplification does come at a cost: the most effective sentence to add might require only a subset of information from some sentence in the document, or incorporate information from more than one sentence.

\paragraph{Sum-Basic (SB):} One common baseline is Sum-Basic, an extractive summarization technique that relies on word frequency statistics to select salient sentences~\cite{nenkova2005impact}.
As an initial step, unigram probabilities are computed from the set of input documents using relative frequency estimation.
Then, sentences are selected one-by-one in greedy rounds until the summary budget is saturated.
At each round, this model selects the most likely sentence according to the current unigram distribution.
The selected sentence is added to the summary and removed from the pool of available sentences.
The unigram probabilities of all words in the selected sentence are heuristically discounted (replaced by square root).
Select-then-discount operations continue until the summary is written.
Discounting is crucial to prevent repetition: once a word (or ideally a concept) has been selected for the summary, it is much less likely to be picked in a subsequent round.

We use Sum-Basic as a Context Agnostic extractive model: we provide the document $\boldsymbol{d}$ as an input to the model and run Sum-Basic for exactly one round.
The selected sentence is considered to be the update sentence $\boldsymbol{x}$. 

\paragraph{Context Informed Sum-Basic (CISB):} We developed a simple modification of the Sum-basic technique to incorporate information from the curated text $\boldsymbol{s}$ as context. 
Initial unigram probabilities are computed using word counts from \emph{both} the curated text \emph{and} the document.
Next, for each sentence in the curated text, we apply just the discount procedure, updating the probability distribution as if those sentences were selected.
Finally, we select the single sentence from the document that is most likely according to the resulting discounted unigram probabilities.
This simple modification of Sum-Basic helps select a sentence that is novel with respect to the curated text by lowering the probability of all words already present.

\paragraph{Extractive CAG, CIG, CRG Models:} 
Any generative model of $\boldsymbol{x}$ can also be used as an extractive model:
we simply estimate the likelihood of each sentence in the document according to the model, and select the most likely one.
Generative models may fail because either they are unable to select the most relevant information, or because the resulting sentence is ill-formed.
Extractive ranking circumvents all errors due to generation and can help isolate model issues.

\paragraph{Hybrid CAG, CIG, CRG Models:}
Since the document $\boldsymbol{d}$ can be quite large, a generative model may struggle to pick the most salient information based on the context.
To simplify the generative modeling task, we can pre-filter the document toward only the most salient parts.
We use the Context Informed Sum-Basic technique to first select the top five sentences from the document.
We supply only these five sentences in place of the source document $\boldsymbol{d}$, then apply the CAG, CIG, and CRG techniques described above.

\section{Dataset}

\begin{figure}[t]
\centering
\includegraphics[scale=0.36]{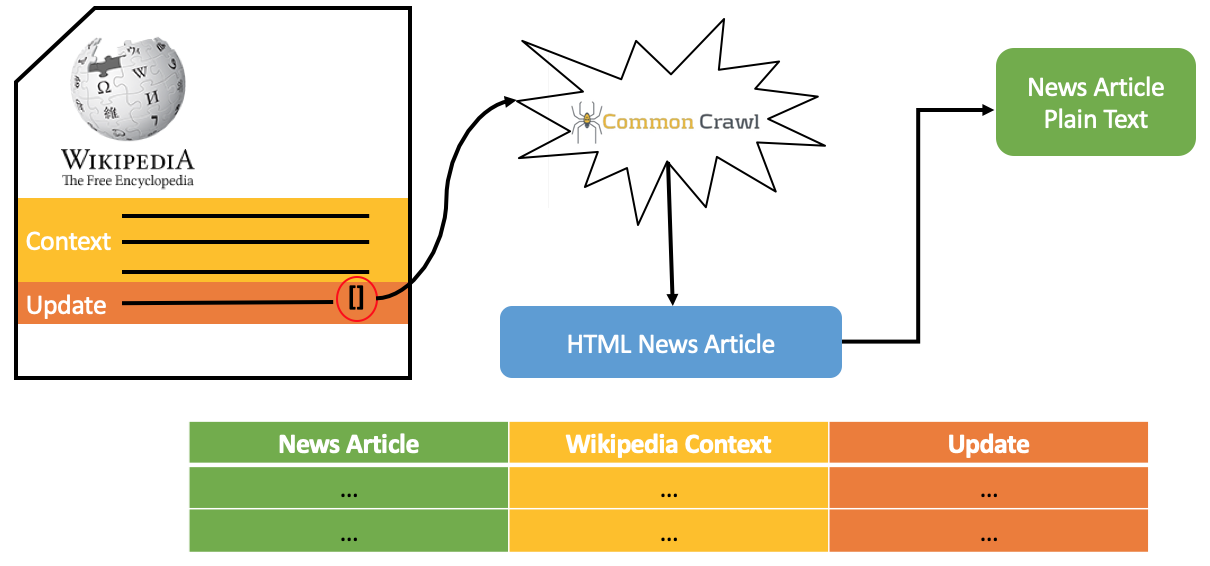}
\caption{Dataset creation process}
\label{fig:data_creation}
\vspace{-1em}
\end{figure}

\begin{table*}[t]
\centering
\begin{tabularx}{\textwidth}{@{}X r r r r@{}}
\toprule
Corpus & Input & Output & \#Examples & Rouge-1 R \\
\midrule
Gigaword \cite{gigaword} & $10^1$ & $10^1$ & $10^6$ & 78.7 \\
CNN/DailyMail \cite{nallapati2016abstractive} & $10^2$--$10^3$ & $10^1$ & $10^5$ & 76.1 \\
WikiSum \cite{liu2018generating} & $10^2$--$10^6$ & $10^1$--$10^3$ & $10^6$  & 59.2  \\
Content Transfer (this paper) & $10^1$--$10^3$ & $10^1$--$10^2$ & $10^5$  & 66.9  \\
\bottomrule    
\end{tabularx}
\caption{Key characteristics of the dataset: approximate size of input and output instances, approximate dataset size, and recall of reference output against the source material, as a measure of dataset difficulty.}
\label{tab:task-corpora-sizes}
\end{table*}

Our ideal dataset would capture the edits made to some curated reference text in light of a stream of new articles describing changes.
For instance, one might maintain reference software documentation about a system, making additions or changes in light of incoming emails describing updates or additions.
This type of data is unfortunately difficult to obtain due to privacy considerations.


However, Wikipedia can provide a naturally-occurring body of text with references to primary sources.
A substantial fraction
of Wikipedia sentences include citations to supporting documentation,
a ripe source of data for content transfer.
That said, some of the citations are quite difficult to follow or trust: broken URLs might lead to lost information; citations to books are difficult to consume given the large scope of information; etc.
Therefore, we only consider cases where the reference links to some well-known news sources.

Based on citation frequency, we selected a list of 86 domains,\footnote{This list is provided in the data release of this paper.} primarily news outlets.
During the data creation process we only considered citations belonging to one of these eighty six domains.
We make this simplifying assumption for several reasons.
First, our English Wikipedia dump contained approximately 23.7 million citation URLS  belonging to 1.6 million domains; fine-grained filtering would be a daunting task.
Our hand-vetted list of domains is a high-precision (albeit low-recall) means of selecting clean data.
Second, we wanted to ground the generated text on credible, consistent, and well-written sources of information. 
Furthermore, well-known domains are readily available on Common Crawl,\footnote{\url{http://commoncrawl.org/}} leading to an easily-reproducible dataset.

Fig.~\ref{fig:data_creation} illustrates the procedure used to create a dataset for the task described in Section~\ref{sec:content_transfer} from Wikipedia. 
For each Wikipedia article, we extracted the plain text without markdown. 
When encountering a citation belonging to a selected domain, we considered the sentence just before the citation to be generated based on the content of the citation.
This sentence became our reference update sentence:
the additional update sentence $\boldsymbol{x}$ added to the curated text $\boldsymbol{s}$ to produce the new text $\boldsymbol{s}'$.
The $k$ sentences prior to the target sentence in the Wikipedia article were considered to be the curated text $\boldsymbol{s}$. 
In our case, we used a window of $k=3$ sentences to select our context.
The cited article acted as the document $\boldsymbol{d}$, from which the appropriate update $\boldsymbol{x}$ can be generated.

The HTML source of the citation was downloaded from Common Crawl for reproducibility and consistency.
The HTML derived from Common Crawl is then processed to get the plain text of the news article.
The resulting dataset $\boldsymbol{C}$ consists of aligned tuples $\boldsymbol{C} = \big(\boldsymbol{d_i}, \boldsymbol{s_i}, \boldsymbol{x_i}\big)_{i \in [1, n]}$, where $n$ is the total number of samples in the dataset.

Alternatively, one might rely on Wikipedia edit history to create a dataset.
In this setting, edits which include a new citation would act as the update $\boldsymbol{x}$.
Although this has the upside of identifying potentially complex, multi-sentence updates, preliminary analysis suggested that these edits are noisy.
Editors may first generate the content in one edit, then add the citation in a subsequent edit, they may only rephrase a part of the text while adding the citation, or they may check in a range of changes across the document in a single edit.
Our simpler sentence-based approach leads to an interesting dataset with fewer complications.
\paragraph{Dataset Statistics and Analysis}
Table~\ref{tab:task-corpora-sizes} describes some key statistics of our dataset and how it compares with other datasets used for similar tasks.
The ROUGE-1 recall scores of reference output $\boldsymbol{x}$ against document $\boldsymbol{d}$ suggest this task will be difficult for conventional extractive summarization techniques.\footnote{ROUGE-1 recall was computed on a sample of 50k instances from the entire dataset.}
%
We hypothesize that during content transfer,
the language in document $\boldsymbol{d}$ often undergoes substantial transformations to fit the curated text $\boldsymbol{s}$.
The average unigram overlap (after stopword removal) between the document $\boldsymbol{d}$ and the reference update sentence $\boldsymbol{x}$ is $55.79$\%; overlap of the curated text $\boldsymbol{s}$ and the reference update sentence $\boldsymbol{x}$ is $30.12$\%.
This suggests the reference update sentence $\boldsymbol{x}$ can be derived from the document $\boldsymbol{d}$, though not extracted directly. 
Furthermore, the content of $\boldsymbol{x}$ is very different from the content of $\boldsymbol{s}$ but appears topically related.

Our dataset consists of approximately 290k unique Wikipedia articles. 
Some heavily-cited articles include `Timeline of investigations into Trump and Russia (2017)', `List of England Test cricketers', and `2013 in science'.
We randomly split the dataset into 580k training instances, 6049 validation instances, and 50k test instances,
ensuring that any Wikipedia article appearing in the train set must not appear in validation or test.

\section{Experimental results}

We evaluate our models using both automated metrics and, for a subset of promising systems, human assessment.
One key evaluation is the similarity between the model generated update sentence and reference update sentence.
We also ask human judges to assess grammaticality and coherence.

\paragraph{Hyper-parameter settings:} For all our experiments with generative models, we have used bidirectional encoder, 2 layers in encoder and decoder, RNN size of 128, word vector size of 100.
We have used sentencepiece toolkit\footnote{https://github.com/google/sentencepiece} to use byte-pair-encoding (BPE) with a vocabulary size of 32k.
We used stochastic gradient descent optimizer and the stopping criterion was perplexity on the validation set.
We filtered our dataset to contain instances which have length of the document between 50 and 2000 tokens, length of the curated text between 20 and 500 tokens and the length of the update sentence between 5 and 200 tokens.

\subsection{Automated Evaluation}

\begin{table}[t]
\centering
\begin{tabularx}{\columnwidth}{@{}X@{}r@{\hspace{0.5em}}lr r @{}}
\toprule
Model & \multicolumn{2}{l}{\rotatebox{0}{ROUGE-L}} & \rotatebox{90}{BLEU} & \rotatebox{90}{METEOR}\\
\midrule
SB                  &  5.6&\confint{5.6--5.7} & 0.6 & 2.0 \\
CISB     &  7.0&\confint{7.0--7.1} & 1.0 & 2.8 \\
\midrule
CAG                        &  9.1&\confint{9.0--9.2} & 1.2 & 4.6 \\
COG                        &  13.5&\confint{13.4--13.6} & 1.7 & 3.5 \\
CIG                        &  \textbf{16.0}&\confint{15.9-16.1} & \textbf{3.5} & \textbf{5.3} \\
CRG                        & 14.7&\confint{14.6--14.8}& 2.6 & 4.5 \\
\midrule
Hybrid CAG                 &  8.0&\confint{7.9--8.0}& 1.0 & 3.8 \\
Hybrid CIG                 & 15.0&\confint{14.9--15.1}& 2.7 & 4.7 \\
Hybrid CRG                 & 13.5&\confint{13.4--13.6}& 2.3 & 4.1 \\
\midrule
Extractive CAG             &  9.3&\confint{9.2--9.3}& 1.1 & 3.2 \\
Extractive CIG             &  9.3&\confint{9.2--9.3}& 1.1 & 3.2 \\
Extractive CRG             &  9.2&\confint{9.1--9.3}& 1.1 & 3.2 \\
\midrule
{\it Oracle} &  {\it 28.8}&{\it \confint{28.7--29.0}} & {\it 11.0} & {\it 10.9}\\
\bottomrule    
\end{tabularx}
\caption{
Automated metrics; 95\% confidence interval in parentheses.
}
\label{tab:rouge_results}
\end{table}

Our primary automated evaluation metric for system-generated update sentences is ROUGE-L F1 against reference update sentence,\footnote{We use the pyrouge toolkit along with ROUGE-1.5.5: \url{https://github.com/bheinzerling/pyrouge}} though we also include BLEU~\cite{papineni2002bleu} and METEOR~\cite{denkowski2011meteor} as additional indicators.
ROUGE is a standard family of metrics for summarization tasks;
ROUGE-L measures the longest common subsequence between the system and the reference, capturing both lexical selection and word order.

Table~\ref{tab:rouge_results} illustrates that this task is quite difficult for extractive techniques.
Furthermore, the results emphasize the importance of having curated text as context when generating the update.
In all experimental conditions, models aware of context perform much better than models agnostic of it.
In contrast to \citet{liu2018generating}, generative approaches outperformed hybrid, likely because we only had a single input document.
Extractive CAG, CIG, and CRG all outperformed both Sum-Basic and the context informed variant.
Extractive CAG was on-par with generative CAG, suggesting the generated sentences were of reasonable quality.
However, generative CIG and CRG were substantially better: rewriting to match context was beneficial.

The {\it Oracle} system of Table~\ref{tab:rouge_results} aims to establish an upper limit attainable by extractive methods, using the following oracle experiment: For each test instance $\big(\boldsymbol{d_i}, \boldsymbol{s_i}, \boldsymbol{x_i}\big)$, we enumerate each extracted sentence $\boldsymbol{e}$ of document $\boldsymbol{d_i}$ and select the one with highest ROUGE-L score as {\it Oracle}'s update sentence $\boldsymbol{\hat{x}_i}$ (i.e., $\boldsymbol{\hat{x}_i} = \arg\max_{\boldsymbol{e} \in \boldsymbol{d_i}} \textrm{ROUGE-L}(\boldsymbol{x_i}, \boldsymbol{e})$).
Note this yields a very optimistic upper bound, as the same ground truth $\boldsymbol{x_i}$ is used both to select an extractive sentence from a large pool of candidates and for final automatic metric scoring.\footnote{Previous work has 
shown that this type of oracle can yield upper bounds that are unrealistically high, and they tend to be above human performance \cite[Table 1]{och:04}. One remedy suggested by Och et al. is a round-robin oracle ensuring that the reference (ground truth) used by the argmax is distinct from that of the final automatic evaluation, but that scheme is only possible with a multi-reference test set.} Nevertheless, these oracle results let us draw two conclusions: (1) They give us better perspective to assess the non-oracle systems, and we believe that their seemingly low automatic evaluation scores are quite reasonable relative to the optimistic upper bound (e.g., CIG’s ROUGE-L’s score is 55\% of the oracle). (2) The oracle results suggest that humans are substantially changing the surface realization as they summarize for Wikipedia, as otherwise the oracle results would be much closer to maximum metric scores (i.e., 100\%). This shows that extractive methods are not enough for this task, justifying our use of generation techniques.


\subsection{Human Evaluations}

For careful evaluation of the performance of the 
most promising configurations (CAG and CIG models) we also asked human judges for quality assessments.
We solicited several types of evaluation, including two relative comparisons between pairs of system outputs and an absolute quality evaluation of individual system outputs.

\paragraph{Close to reference (Relative):} The first relative comparison measured how accurately the generated update reflected information in the reference update.
Here, the annotators saw only the reference update sentence and the outputs of two systems labeled $A$ and $B$ in a randomized order.
We asked the annotators ``Which system output is closest in meaning to the reference update?''
The annotators could pick system $A$, system $B$, or indicate that neither was preferred.
This is a simple evaluation task though potentially biased toward the sole reference update.

\paragraph{Coherent to context (Relative):} The second relative comparison measured whether the generated output contained salient information from the document written in a manner appropriate to the curated text.
The annotators saw the document $\boldsymbol{d}$, the curated text $\boldsymbol{s}$, and the outputs of the two systems $A$ and $B$, again in a random order.
They were asked, ``Which system output is more accurate relative to the background information given in the snippet of the article?''
Each judge had to consider whether the information fits with the curated text and also whether system-generated content could be supported by the document.

Four human judges each annotated 30 unique output pairs for these two relative comparison settings, a total of 240 relative judgments.
Table~\ref{tab:human_rel_results} shows the results: the context-aware CIG system was substantially better in both settings.

\paragraph{DUC Guidelines (Absolute):} In addition, we performed an absolute quality evaluation following the guidelines from DUC 2007.\footnote{\url{http://duc.nist.gov/duc2007/quality-questions.txt}}
Each judge was presented with a single system output, then they were asked to evaluate five aspects of system output: grammaticality, non-redundancy, referential clarity, focus, and structure/coherence.
For each aspect, the judge provided an assessment on a five-point scale: (1) Very Poor, (2) Poor, (3) Barely Acceptable, (4) Good, (5) Very Good.
We gathered 120 additional judgments in this setting (4 judges, 30 outputs).
Again, context-aware CIG substantially outperforms CAG across the board, as seen in Table~\ref{tab:human_abs_results}.

\begin{table}[t]
\centering
\begin{tabularx}{\columnwidth}{@{}X r r r @{}}
\toprule
& \multicolumn{3}{c}{prefer} \\
\cmidrule{2-4}
Evaluation task & CAG & neither & CIG \\
\midrule
Close to reference      & 15.8\% & 53.3\% & 30.8\% \\ 
Coherent to context       & 7.5\% & 53.3\% & 39.2\% \\
\bottomrule    
\end{tabularx}
\caption{
Human preferences of CAG vs. CIG.
}
\label{tab:human_rel_results}
\end{table}

\begin{table}[t]
\centering
\begin{tabularx}{\columnwidth}{@{}X r r r r r@{}}
\toprule
Model &
\rotatebox{90}{Grammar} &
\rotatebox{90}{Non-redund.} &
\rotatebox{90}{Ref. clarity} &
\rotatebox{90}{Focus} &
\rotatebox{90}{Structure}\\
\midrule
CAG & 2.6 & 1.8 & 2.7 & 2.6 & 2.4 \\
CIG & \textbf{4.3} & \textbf{3.9} & \textbf{3.6} & \textbf{3.5} & \textbf{3.2} \\
\bottomrule    
\end{tabularx}
\caption{
Human absolute quality assessments.
}
\label{tab:human_abs_results}
\end{table}

\begin{figure}[t]
\centering
\footnotesize
\begin{tabularx}{\columnwidth}{@{}X@{}}
\toprule
\tabsec{Document (News Article)}
sequels are fairly new to bollywood, but director sanjay gadhvi realised there was cash to be made from resurrecting his hit action thriller dhoom, by casting sexy young stars like hrithik rosha, aishwarya rai and abhishek bachchan in an even bigger game of cops and robbes...that the twist in dhoom 2's tail is not explained is yet another shortcoming.
\hlArt{it's only roshan's charismatic performance as the criminal mastermind, and the sizzling chemistry he shares with rai's sassy cohort, that rescues this adventure from becoming an elongated tourism commercial.}
\\
\midrule
\tabsec{Curated Text (Wikipedia Context)}
\hlCtx{it makes no lasting contributions to world cinema, but if two-and-a-half hours of disposable entertainment are all you're after, you could do far worse. ``l.a. weekly's david chute stated the film was, "a movie meal as satisfying as this one can make you feel that nothing else matters.'' jaspreet pandohar of the bbc gave it a two-star rating, writing ``by roping in acclaimed action director alan amin to take care of the thrills and spills, you'd expect gadhvi to have spent time crafting out a sophisticated storyline instead of simply sending his cast on a cat-and-mouse chase around the globe.} \\
\midrule
\tabsec{Reference Update}
\hlRef{it's only roshan's charismatic performance as the criminal mastermind, and the sizzling chemistry he shares with rai's sassy cohort, that rescues this adventure from becoming an elongated tourism commercial."}
\\
\midrule
\tabsec{Generated Update}
it's only roshan's finest performance as the criminal terrorist, and the sizzling chemistry he shares with rai's sassy anatomy, that attues this adventure from becoming an elongated tourism commercial."
\\
\bottomrule
\end{tabularx}
\caption{Example of good quality generation, where the system-generated update is close to the reference.}
\label{fig:generated_example1}
\vspace{-1em}
\end{figure}

\paragraph{Observations:} Systems unaware of the curated text $\boldsymbol{s}$ tend to generate long updates with repeated frequent words or phrases. 
Consider the ratio of unique tokens over the total number of tokens in the generated output, which we denote by $\boldsymbol{R}$. 
A small $\boldsymbol{R}$ indicates many repeated tokens.
We find that $88$\% of the time this ratio $\boldsymbol{R}$ falls below $0.5$ for the CAG model, i.e. for $88$\% instances, more than $50$\% of the words in the generated output are repeats.
This number is relatively small -- $14$\% for CIG and $20$\% for CRG -- in context aware models.
In the reference updates only $0.21$\% instances repeat more than $50$\% of words.

Figs.~\ref{fig:generated_example1} and \ref{fig:generated_example2} show good and bad examples generated by the CIG model along with the document, curated text and the reference update.
Table~\ref{tab:gen_ex} has a set of updates generated by the CIG model as well as the reference update.
As we can see in examples 3 and 4, the CIG model misplaces the date but correctly generates the remaining content.
In examples 1 and 2, the CIG model appears to successfully select the correct pronouns for co-reference resolution, though it gets confused as to when to use the pronoun or the named entity.
Examples 5 and 6 represent failure cases due to missing words.




\begin{figure}[t]
\centering
\footnotesize
\begin{tabularx}{\columnwidth}{@{}X@{}}
\toprule
\tabsec{Document (News Article)}
anne kirkbride, who portrayed bespectacled, gravelly-voiced deirdre barlow in coronation street for more that four decades, \hlArt{has died. the 60-year-old,} whose first appearance in the soap opera was in 1972, died in \hlArt{a manchester hospital} after a short illness.... kirkbride had left the soap opera after \hlArt{she was diagnosed with non-hodgkin's lymphoma} in 1993 but returned some months later after treatment and spoke candidly about how she had struggled with depression following the diagnosis...
\\
\midrule
\tabsec{Curated Text (Wikipedia Context)}
\hlCtx{in 1993, kirkbride was diagnosis with non-hodgkin's lymphoma. she spoke to the british press about her bout of depression following the diagnosis. she was cured within a year of being diagnosed.} \\
\midrule
\tabsec{Reference Update}
\hlRef{anne kirkbride died of breast cancer in a manchester hospital on 19 january 2015, aged 60.}
\\
\midrule
\tabsec{Generated Update}
she was diagnosed with non-hodgkin's lymphoma.
\\
\bottomrule
\end{tabularx}
\caption{
Example of lower-quality output: the generated update unnecessarily restates information yet misses the most salient detail from the document.}
\label{fig:generated_example2}
\vspace{-1em}
\end{figure}

\begin{table*}[t]
\centering
\small
\begin{tabularx}{\textwidth}{@{}l@{\hspace{0.5em}}XX@{}}
\toprule
& Reference Update & Generated Update\\
\midrule
1. & rob brydon, the comedian was born in baglan.                                                 & he was born in baglan.\\
2. & in may 2014 he was diagnosed with prostate cancer.                                           & st. clair was diagnosed with prostate cancer. \\
3. & on april 3, 2014, manning signed a one-year deal with the cincinnati bengals. & on march 9, 2014, manning signed a one-year contract with the cincinnati bengals. \\
4. & on oct 10, 2013, barrett signed with the memphis grizzlies. & on feb 9, 2013, barrett signed with the memphis grizzlies.\\
5. & some people think elvis is still alive, but most of us think he's dead and gone." & some people think elvis, but most of us think he's dead and gone." \\ 
6. & it's always the goal of the foreign-language film award executive committee to be as inclusive as possible." & it's always the goal of the foreign- entry film award executive to be as possible."\\
\bottomrule
\end{tabularx}
\caption{Example generations from the CIG system, paired with the human generated updates.}
\label{tab:gen_ex}
\end{table*}

\section{Related Work}
\label{sec:related}

 
The proposed content transfer task is clearly related to a long series of papers in summarization, including recent work with neural techniques~\cite{rush-chopra-weston:2015,nallapati2016abstractive}.
In particular, one recent paper casts the the task of generating an entire Wikipedia article as a multi-document summarization problem~\cite{liu2018generating}.
Their best-performing configuration was a two-stage extractive-abstractive framework; a multi-stage approach helped circumvent the difficulties of purely abstractive methods given quite large input token sequences.


Looking beyond the clear task similarity of authoring Wikipedia style content, there are several crucial differences in our approach.
First, the goal of that paper is to author the whole page, starting from nothing more than a set of primary sources, such as news articles.
In practice, however, Wikipedia articles often contain information outside these primary sources, including common sense knowledge, framing statements to set the article in context, and inferences made from those primary sources.
Our task restricts the focus to content where a human editor explicitly decided to cite some external source.
Hence, it is much more likely that the resulting summary can be derived from the external source content.
Furthermore, we focus on the act of adding information to existing articles, rather than writing a complete article without any context.
These two scenarios are clearly useful yet complementary: sometimes people want to produce a new reference text where nothing existed before; in other cases the goal is to maintain and update an existing reference.


Another closely related task is update summarization~\cite{dang2008overview}, where systems attempt to provide a brief summary of the novel information in a new article assuming the user has read a known set of prior documents.
Our focus on curating an authoritative resource is a substantial difference.
Also our datasets are substantially larger, enabling generative models to be used in this space, where prior update summarization techniques have been primarily extractive~\cite{fisher2008query,li2015improving}.



For any generation task, it is important to address both the content (`what' is being said) as well its style (`how' it is being said).
Recently, a great deal of research has focused on the `how'~\cite{HeHe, shen2017style}, including efforts to collect a parallel dataset that differs in formality \cite{rao2018dear}, to control author characteristics in the generated sentences \cite{P18-1080}, to control the perceived personality traits of dialog responses \cite{zhang2018personalizing}. 
We believe this research thread is complementary to our efforts on generating the `what'.

Another form of content transfer bridges across modalities:
text generation given schematized or semi-structured information.
Recent research has addressed neural natural language generation techniques given a range of structured sources: selecting relevant database records and generating natural language descriptions of them~\cite{mei2016talk}, selecting and describing slot-value pairs for task-specific dialog response generation~\cite{sem_cond_lstm}, and even generating Wikipedia biography abstracts given Infobox information~\cite{wikibio}.
Our task, while grounded in external content, is different in that it leverages \emph{linguistic} grounding as well as prior text context when generating text.
This challenging setting enables a huge range of grounded generation tasks: there are vast amounts of unstructured textual data.
 

\section{Conclusions}

This article highlights the importance of the task of \emph{content transfer}: generation guided by an existing curated text to set context and tone, and grounded in a new source providing useful information.
We demonstrate how multiple models can address this challenging problem on a novel dataset derived from Wikipedia and Common Crawl.
This dataset is released to the community along with scripts and models.\footnote{\url{https://www.microsoft.com/en-us/research/project/content-transfer/}}
We find this setting particularly promising given the opportunity for human interaction: in contrast to approaches that do not rely on human-generated context, we establish a collaboration between user and computer.
Each newly suggested sentence can be rejected, accepted, or edited before inclusion, and the edits can provide more training data.

We believe there are many natural extensions to this work.
The models described here are mostly extensions of existing approaches;
approaches targeting novelty detection, focus, and document structure could lead to substantial improvements.
We could apply models in series to incorporate changes for a set of documents.
Future work could also explore changes that modify existing content rather than simply appending. 

%

%
 %
%
 

\section*{Acknowledgments}
We are grateful to the anonymous reviewers, as well as
Alan W Black,
Chris Brockett,
Bill Dolan, 
Sujay Jauhar, 
Michael Gamon, 
Jianfeng Gao, 
Dheeraj Rajagopal,
and Xuchao Zhang
for their helpful comments and suggestions on this work. We also thank 
Emily Ahn, 
Khyati Chandu, 
Ankush Das, 
Priyank Lathwal, 
and Dheeraj Rajagopal for their help with the human evaluation.

\bibliographystyle{acl_natbib}
\bibliography{naaclhlt2019}

\end{document}